\documentclass[10pt,twocolumn,letterpaper]{article}

\usepackage[review]{cvpr}      

\usepackage[dvipsnames]{xcolor}

\definecolor{cvprblue}{rgb}{0.21,0.49,0.74}
\usepackage[pagebackref,breaklinks,colorlinks,citecolor=cvprblue]{hyperref}

\usepackage{array}
\usepackage{hyperref}
\usepackage{url}
\usepackage{hyperref}

\usepackage[dvipsnames]{xcolor}
\definecolor{c3}{HTML}{D8ECD1}
\definecolor{c2}{HTML}{A8C9EA}
\definecolor{c1}{HTML}{E4E4E6}
\definecolor{c4}{HTML}{7CCD7C}
\definecolor{c5}{HTML}{F0988C}

\newcommand{\hi}[1]{\textcolor{c4}{#1}}
\newcommand{\hie}[1]{\textcolor{c5}{#1}}

\usepackage[accsupp]{axessibility}  %

\usepackage{graphicx}
\usepackage{amsmath}
\usepackage{amssymb}
\usepackage{booktabs}
\usepackage{multibib}
\usepackage{float}

\usepackage{subcaption}
\usepackage{multirow}
\usepackage{url}
\usepackage{wrapfig}
\usepackage{wrapfig}
\usepackage{diagbox}
\newcommand{\argtopk}{\mathop{\mathrm{arg\,top}\,k}}

\newlength\savewidth\newcommand\shline{\noalign{\global\savewidth\arrayrulewidth
  \global\arrayrulewidth 1pt}\hline\noalign{\global\arrayrulewidth\savewidth}}
\newcommand{\tablestyle}[2]{\setlength{\tabcolsep}{#1}\renewcommand{\arraystretch}{#2}\centering\footnotesize}

\newcolumntype{*}{>{\global\let\currentrowstyle\relax}}
\newcolumntype{^}{>{\currentrowstyle}}

\title{POS: A Prompts Optimization Suite for Augmenting Text-to-Video Generation}

\author{First Author\\
Institution1\\
Institution1 address\\
{\tt\small firstauthor@i1.org}
\and
Second Author\\
Institution2\\
First line of institution2 address\\
{\tt\small secondauthor@i2.org}
}

\begin{document}
\maketitle
\begin{abstract}
This paper targets to enhance the diffusion-based text-to-video generation by improving the two input prompts, including the noise and the text. Accommodated with this goal,  we propose POS, a \textbf{P}rompt \textbf{O}ptimization \textbf{S}uite to boost text-to-video models.
POS is motivated by two observations: \textbf{(1) Video generation shows instability in terms of noise.} Given the same text, different noises lead to videos that differ significantly in terms of both frame quality and temporal consistency. This observation implies that there exists an optimal noise matched to each textual input; 
To capture the potential noise, we propose an optimal noise approximator to approach the potential optimal noise. Particularly, the optimal noise approximator initially searches a video that closely relates to the text prompt and then inverts it into the noise space to serve as an improved noise prompt for the textual input.
\textbf{(2) Improving the text prompt via LLMs often causes semantic deviation.} 
Many existing text-to-vision works have utilized LLMs to improve the text prompts for generation enhancement.
However, existing methods often neglect the semantic alignment between the original text and the rewritten one.
In response to this issue, we design a semantic-preserving rewriter to
impose constraints in both rewriting and denoising phrases to preserve semantic consistency.
Extensive experiments on popular benchmarks show that our POS can improve the text-to-video models with a clear margin. The code will be open-sourced. 
\end{abstract}
  
\section{Introduction}
Text-to-video generation has emerged as a valuable approach in automated video production, offering a human-friendly method with wide applications in diverse industries such as media, gaming, and film. Recently, diffusion-based text-to-video generation has witnessed remarkable strides~\cite{ho2022video,singer2022make,ho2022imagen,zhou2022magicvideo}. Following the paradigm of conditional diffusion models \cite{rombach2021highresolution},
the text-to-video generation models typically sample a Gaussian noise and a text condition to synthesize the videos. However, during our practice, we found that the noise and the text prompt seriously affected the video quality of existing text-to-video models. 

\begin{figure}
    \centering
    \includegraphics[width=8cm]{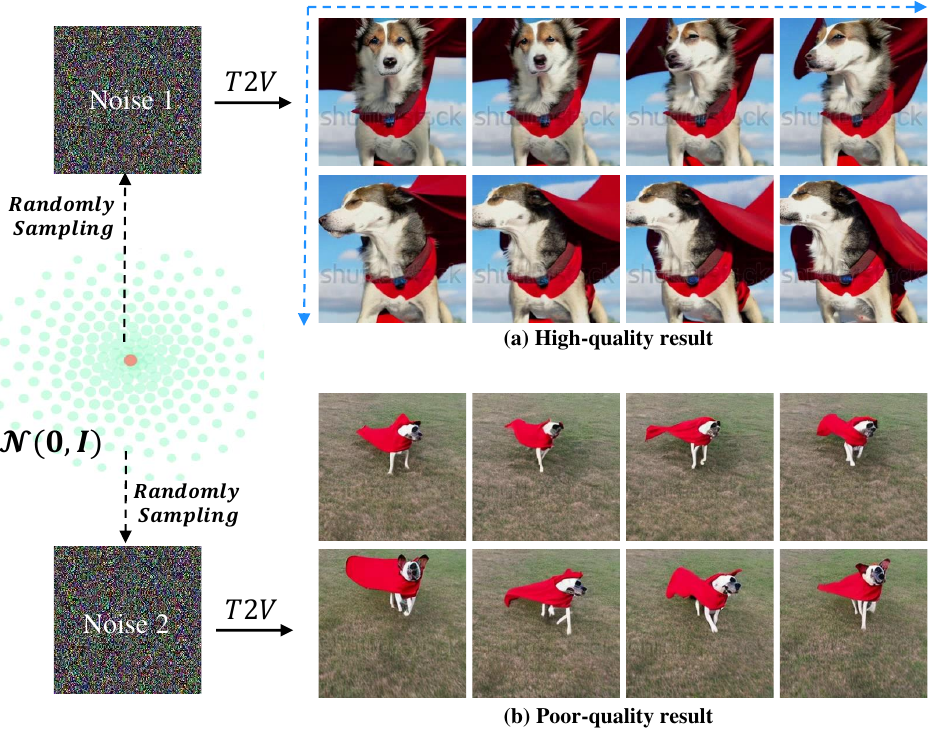}
    \vspace{-0.2cm}
    \caption{Different noises can yield significantly different videos in terms of quality. With this observation, we posit there exists a potential optimal noise (orange circle), randomly sampled noise close to the optimal noise can synthesize high-quality results, while the noise far away leads to poor quality.
    Both videos are produced by ModelScope~\citep{wang2023modelscope} with the same prompt ``\emph{A dog wearing a Superhero outfit with red cape flying through the sky}''. }
    \label{fig:noiseisimportant}
\end{figure}

\begin{figure*}
    \centering
    \includegraphics[height=5cm]{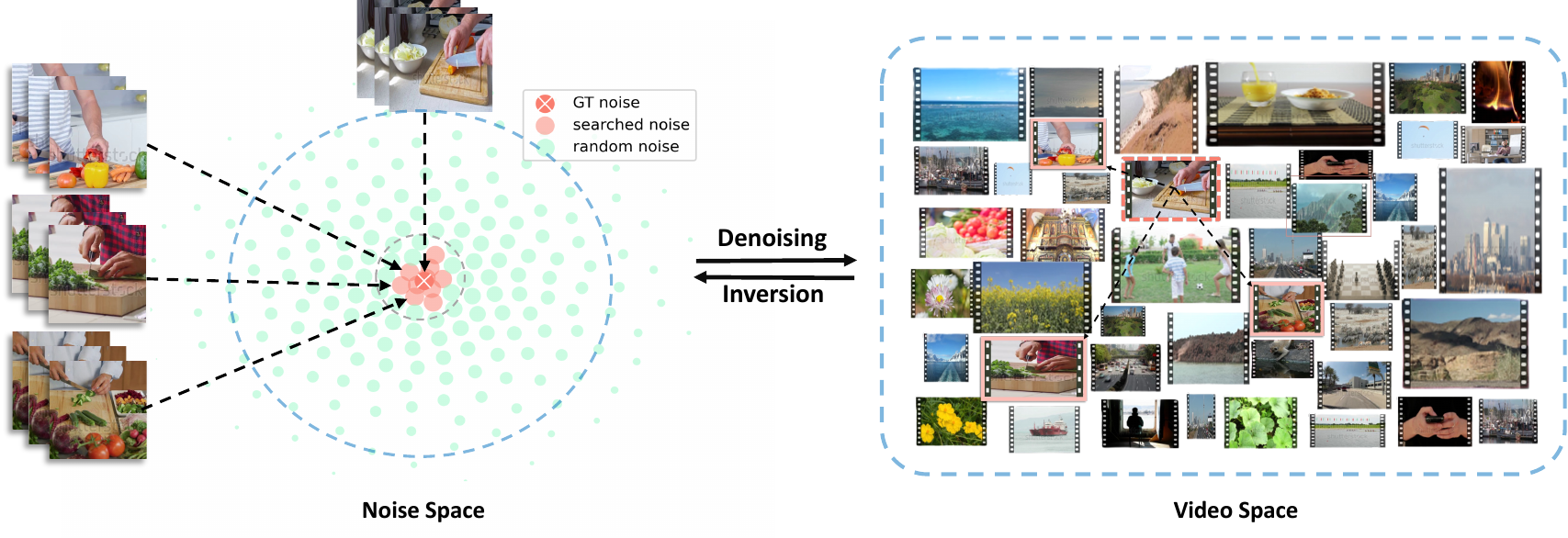}
    \vspace{-0.3cm}
    \caption{Motivation illustration of optimal noise approximator. The trained denoising and inversion functions establish a bidirectional mapping between the video space and the noise space. Treating the inversion of the groundtruth video (``GT noise") as the optimal noise, our objective is to approximate this optimal noise by inverting video neighbors. It is observed that similar videos converge to a confined region within the noise space, forming the theoretical basis for our optimal noise approximator. }
    \label{fig:motivation}
\end{figure*}

Different noises can yield significantly varied videos in terms of frame quality and temporal consistency, just as shown in Figure~\ref{fig:noiseisimportant}. 
This observation highlights the existence of an optimal noise sample for a given text prompt, but the strategy of random sampling noise is hard to hit or get close to it every time. As a result, video quality varies, with higher quality achieved when the noise sample aligns closely to the optimal point, and poorer quality observed when it deviates further away.
Given this consideration,  we target to approach the optimal noise for a given text prompt to consistently generate high-quality videos.
Notably, the inversion and denoising procedures within the diffusion model establish a bidirectional mapping between the noise space and the video space. Based on the fact that the inversion of a sample can almost reconstruct itself through denoising~\cite{mokady2023null, lu2023tficon, wallace2022edict}, we think that the inversion of the groundtruth video of the given text prompt is close enough to the optimal noise.
Nevertheless, the groundtruth video is not available during inference. 
To tackle this challenge, we propose to leverage the neighboring counterpart of the desired video to approximate the optimal noise.

To be specific, we first search a neighbor video for the text input, and then apply the inversion procedure on it to locate a point in the noise space as shown in Figure~\ref{fig:motivation}.
Motivated by this concept, we initially assemble a pool of text-video pairs. For a given input text prompt, we select the video from the pair whose text is similar to the text prompt as the neighbor video. Subsequently, this chosen video is inverted into the noise space as an approximation for the optimal noise. Finally, the approximated noise is input to the forward procedure in the diffusion model, resulting in the synthesis of higher-quality videos. To get rid of the storage pressure of the retrieval pool, we also take the text-inversion noise pairs as the training samples, and train an optimal noise prediction network to output the optimal noise directly. 
However, we note that the fixed noise for every text would lead to poor diversity, to remedy this issue, we further augment the noise initialization by introducing the random noise via Gaussian Mixture to maintain the diversity. 

In addition to noise inputs, text prompts also play a pivotal role in influencing the quality of generated videos. Detailed descriptions tend to produce superior results, and many works have utilized large-scale language models (LLMs) to enhance the text prompts \cite{zhu2023moviefactory,hao2022optimizing, hong2023large,openai2023dalle3}.  For the text-to-video generation, we can straightforwardly enhance the text descriptions using LLMs such as ChatGPT \cite{openai2023gpt4} and Llama2 \cite{touvron2023llama}. However,this na\"ive strategy presents two issues: 
\textbf{(a)} The potential information gained through simple rewriting is often limited, resulting in incremental improvements to the text prompts.
\textbf{(b)} Without appropriate guidance, the introduction of unexpected content may lead to the generation of videos that deviate from the user's original intention. In light of these issues, we propose a semantic-preserving rewriter, formed by a Reference-Guided Rewriting mechanism and Denoising with Hybrid-Semantics to respectively remedy the above two challenges. 

Particularly, Reference-Guided Rewriting (RGR) searches several texts as the reference for the rewriting, serving as the information pool to aid the LLMs in compensating reasonable details for input text.
To preserve the semantics, we design a  Denoising with a Hybrid-Semantics (DHS) strategy, the rewritten text acts as the contextual condition in the early diffusion steps for quality enhancement, while the original text is introduced in later steps of the denoising process to align the semantics with the original text. By strategically including the original text, DHS ensures that the final video closely adheres to the original text, maintaining the intended narrative consistency. 
Through advancements in enhancing the two inputs of text-to-video models, we have successfully developed an optimization-free and (diffusion) model-agnostic method. 

In summary, we contribute a prompts optimization suite (POS) for text-to-video generation, featured by two components: \textbf{1) An optimal noise approximator} (ONA) that offers the noise initialization close to the optimal noise for each text prompt. \textbf{2) A semantic-preserving rewriter} (SPR) characterized by the RGR and DHS to provide detail-rich text prompts while maintaining the semantics of the final video to keep consistent with the original text.

\begin{figure*}
    \centering
    \includegraphics[height=5.6cm]{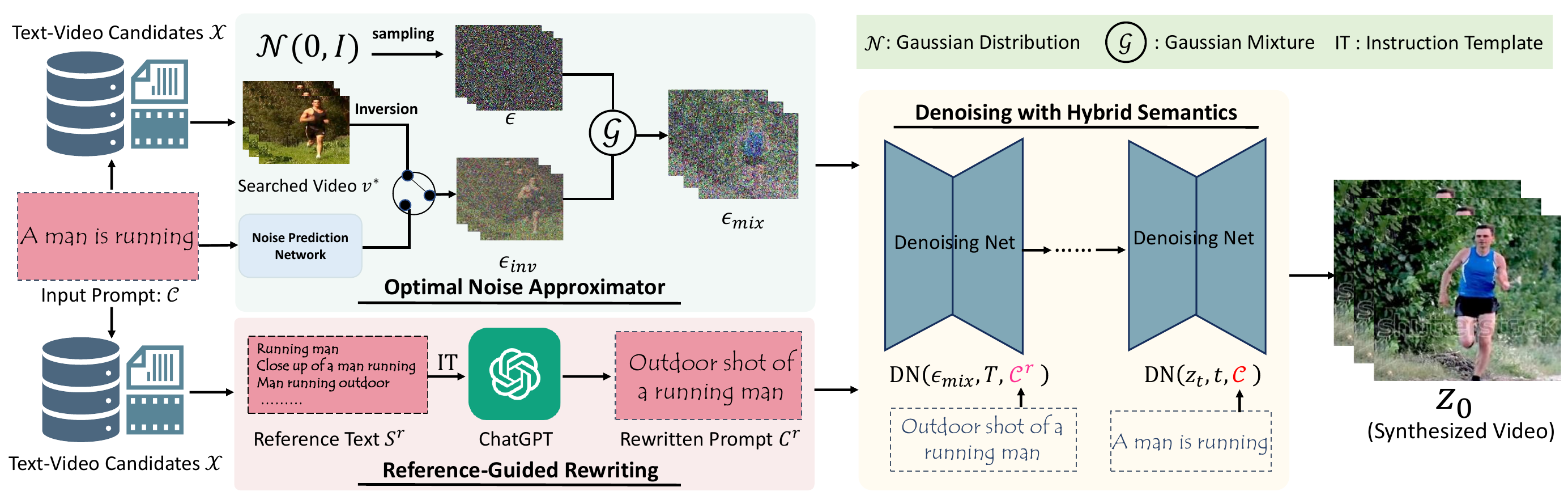}
    \vspace{-0.2cm}
    \caption{Illustration of our POS. Given a trained text-to-video model, POS enhances it by improving the two types of prompts: the noise and the text. The optimal noise approximator targets to approach the optimal noise for the text prompt, while the semantic-preserving rewriter, formed by the reference-guided rewriting and the denoising with hybrid semantics, improves the text prompt by providing more details without deviating from the original semantics.  }
    \label{fig:method}
\end{figure*}

\section{Related work}
\label{gen_inst}
Text-to-video generation within open domains, using diffusion models, is an active and compelling research area. The prevailing approaches \citep{ho2022video,singer2022make,ho2022imagen,zhou2022magicvideo,blattmann2023align,wu2022tune,ge2023preserve,luo2023videofusion, wang2023videofactory} typically extend text-to-image models to facilitate video generation. This strategy leverages the knowledge from text-to-image models. Broadly, existing methods enhance text-to-video generation through two primary avenues: architecture-oriented optimization \citep{ho2022video, singer2022make, ho2022imagen} and noise-oriented optimization \citep{luo2023videofusion, ge2023preserve}. Architecture-oriented optimization typically designs modules for temporal relation enhancement. For example, 
VideoFactory \citep{wang2023videofactory} proposes a novel spatiotemporal cross-attention to reinforce the interaction between spatial and temporal features. 
While noise-oriented optimization seeks to provide precise noise initialization to maintain video fluency more effectively. 
VideoFusion \citep{luo2023videofusion} and Preserve Your Own Correlation~\cite{ge2023preserve} both target to maintain the frame coherence from the noise input.
Instead of designing the noise empirically,  this work targets to directly approximate the optimal noise for the text prompt. What's more, we also enrich the text prompts via the LLMs. Notably, the proposed POS is optimization-free and model-agnostic, which are also important features that distinguish our method from existing models.
\section{Preliminaries}
Before diving into the details of our method, we first revisit two crucial modules in Denoising Diffusion Implicit Models (DDIMs) \cite{song2020denoising}, \emph{i.e.,} the Denoising and the Inversion procedures,
to ease the understanding of our method.
\noindent\textbf{Denoising} in diffusion model builds a mapping from normal noise to samples (image, video \emph{etc}). Compared with Denoising Diffusion Probabilistic Models (DDPMs) \cite{ho2020denoising}, DDIMs \cite{song2020denoising} are more efficient, allowing for the use of a smaller number of steps to synthesize a sample $z_0$ from a Gaussian noise $z_T$ using the iterative procedure:
\begin{equation}
\label{denoising}
    z_{t-1}=\sqrt{\alpha_{t-1}}(\frac{z_{t}}{\sqrt{\alpha_{t}}}+(\sqrt{\frac{1-\alpha_{t-1}}{\alpha_{t-1}}}-\sqrt{\frac{1-\alpha_{t}}{\alpha_{t}}})\epsilon_{\theta}(z_{t},t,c)),
\end{equation}
where $\alpha_{t} = \prod_{t=1}^T(1-\beta_{t})$, $\beta_{t}$ is a predefined hyperparameter in DDPMs, $c$ is the condition to control the generation, $\epsilon_{\theta}$ is a noise prediction network. We define function $\text{DN}(z_{t},t,c)$ as the right term of the above equation to simplify the subsequent description: $z_{t-1} = \text{DN}(z_{t},t,c)$. 

\noindent\textbf{Inversion} defines a projection from data space to noise space. Given a sufficiently large $T$, Eq.\ref{denoising} approaches an ordinary differential equation (ODE). With the assumption that the ODE process can be reversed in the limit of small steps \cite{mokady2023null}, we can acquire a noise from a data $z_0$:
\begin{equation}
    z_{t+1}=\sqrt{\alpha_{t+1}} (\frac{z_{t}}{\sqrt{\alpha_{t}}}+(\sqrt{\frac{1-\alpha_{t+1}}{\alpha_{t+1}}}-\sqrt{\frac{1-\alpha_{t}}{\alpha_{t}}})\epsilon_{\theta}(z_{t},t,c)).
\end{equation}

Analogously, we define the right term of the above equation as $\text{INV}(z_{t},t,c)$ to convince the subsequent elaboration: $z_{t+1} = \text{INV}(z_{t},t,c)$. With the above preliminaries, we next elaborate on the key components of our POS, \emph{i.e.,} optimal noise approximator and semantic-preserving rewriter.

\section{Methodology}
\label{headings}
Figure~\ref{fig:method} depicts our framework,  POS, formed by ONA and SPR, which can be seamlessly integrated into the existing text-to-video generation frameworks. Given a trained text-to-video model and the text prompt, ONA produces an improved noise by video inversion or directly generating with the noise prediction network, which is then combined with random noise to form the final noise prompt. SPR locates textual references for LLMs rewriting and performs denoising with hybrid semantics for semantics preservation.
\subsection{Optimal Noise Approximator}
\label{ona}
\noindent\textbf{Video Retrieval} seeks to find a video $v*$ that aligns with the text prompt in semantics. We first prepare a pool of $N$ text-video pairs $\mathcal{X}=\{S_i, V_i\}_{i=1}^{N}$, where $S$ and $V$ represent text and video, respectively. Given a text prompt $\mathcal{C}$, we initially estimate the similarity between the text prompt and the text within $\mathcal{X}$ and then select the video accordingly:
\begin{equation} \label{eq6}
    v* = \{V_{i}|\arg\max \limits_{i}\{\text{sim}(E_t(\mathcal{C}),E_t(S_{i}))|\{S_{i},V_{i}\}\in \mathcal{X}\}\}
\end{equation}
where $\text{sim}(\cdot,\cdot)$ is the cosine similarity, and $E_t$  can be a off-the-shelf  language model like \citep{reimers-2019-sentence-bert, devlin2019bert}, to extract the text feature. The reasons we anchor on the text-similarity for the video selection are three-fold: first, text similarity is a more reliable clue than the text-video relevance; second, the computation efficiency for text feature extraction is superior; third, due to the available text-video dataset like WebVid-10M \citep{bain2021frozen}, InvernVid \citep{wang2023internvid}, collecting the text-video pairs is not effort-intensive as well.   

\noindent\textbf{Guided Noise Inversion} module takes the searched $v*$ as the inversion source to approximate the optimal noise.
Formally, we follow the efficient latent diffusion architecture \cite{rombach2021highresolution}, where the raw video $v*$ is first encoded via the encoder of VQ-VAE \cite{razavi2019generating} and then the inversion is recurrently acted on the latent feature with $T$ steps: $\epsilon_{\text{inv}} = \text{INV}(\mathcal{E}(v*),t,\varnothing)|_{t=0}^{T-1}$, where $\epsilon_{\text{inv}}$ is the inverted noise, $\varnothing$ represents the empty text and $\mathcal{E}$ is the VQ-VAE encoder. With the video as the inversion source, $\epsilon_{\text{inv}}$ can also inherit the coherence of $v*$, thereby benefiting the temporal consistency as well.

As shown in Figure~\ref{fig:motivation}, this simple strategy can help locate a point close to the optimal noise. On the other hand, the prompts that share the inversion source video will use the same noise; especially, multiple inferences of the same prompt will yield the same result, severely sacrificing the diversity.
In response to this shortcoming, we introduce the Gaussian noise and integrate it with the inverted noise $\epsilon_{\text{inv}}$ via Gaussian mixture to maintain the randomness:
\begin{equation} \label{eq7}
    \epsilon_\text{mix}(v*, \eta) = \frac{1}{\sqrt{1+\eta^2}}\cdot \epsilon + \frac{\eta}{\sqrt{1+\eta^2}}\cdot \epsilon_{\text{inv}},
\end{equation}
where $\epsilon \sim \mathcal N(0,\mathbf{I})$, $\eta$ is the hyperparameter to balance the noises. Noise mixture in Eq.~\ref{eq7} is inspired by PYoCo\cite{ge2023preserve}. Unlike PYoCo, which performs a mixture between noise frames to achieve smoothness, the noise mixture here is a tool to introduce random noise for diversity preservation, which is not the paper's focus.



\noindent\textbf{Video Synthesize with Improved Noise.} 
Given the improved noise $\epsilon_\text{mix}$ and text prompt $\mathcal{C}$, the latent code $z_{0}$ of the video is calculated by the following formula:
\begin{equation}
\label{inference}
    z_{t-1}=\left\{
    \begin{array}{lcl}
    \text{DN}(\epsilon_\text{mix}(v*,\eta),t,\mathcal{C}) & & \text{if}\;{t=T}\\
    \text{DN}(z_{t},t,\mathcal{C}) & & \text{if}\;{0 < t \leq T-1 },
    \end{array} \right.
\end{equation}
where $T$ is the total DDIM sampling steps.

\subsection{Noise Prediction Network}
\begin{figure*}[t]
    \centering
    \includegraphics[height=7.2cm]{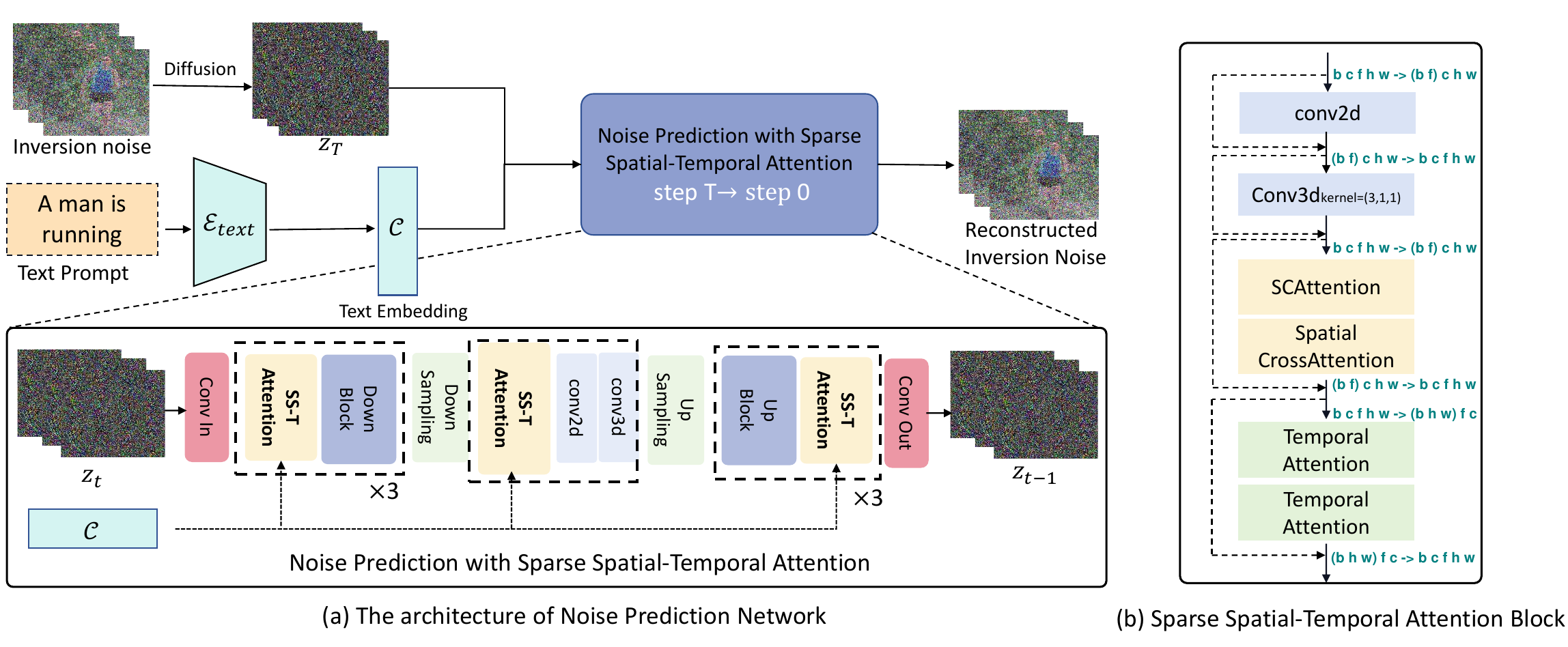}
    \vspace{0.01cm}
    \caption{Architecture of Noise Prediction Network. During training, we invert real videos into the noise space, pairing them with their corresponding text descriptions to form training pairs. At inference time, the text prompt is directly input into the trained network to yield the optimal noise prediction.}
    \label{fig:NPNet}
    \vspace{-0.5cm}
\end{figure*}

Following subsection~\ref{ona}, we can acquire $\epsilon_\text{inv}$ in a training-free manner. However, it requires a paired video-text pool as the prerequisite. To circumvent this limitation, we train a Noise Prediction Network (NPNet) to generate the optimal noise directly.
As illustrated in Figure ~\ref{fig:NPNet}, NPNet follows the video diffusion paradigm.  
It comprises two parts: 1) Text encoder CLIP \cite{radford2021learning} extracts features from the input text as conditions. 2) Sparse Spatial-Temporal attention network for denoising, which is built upon the Sparse Spatial-Temporal (SS-T) Attention. Figure~\ref{fig:NPNet}
(b) shows the detailed architecture of SS-T attention, The input feature undergoes processing through consecutive layers including a 2D convolution (conv2D), a 3D convolution (conv3D), Sparse Causal Attention (SCAttn) \cite{wu2022tune}, spatial cross attention, and two temporal attention modules. During training, we leverage inversion noises and paired texts as the training data, and train NPNet using the noise reconstruction loss \cite{ho2020denoising}. Post-training, we obtain the optimal noise by feeding the text prompt through NPNet, which replaces $\epsilon_\text{inv}$ in \cref{eq7}.

NPNet is primarily intended as an alternative to bypass the text-video search pool in our standard ONA framework; however, its architectural design is not the central focus of this study. Therefore, we omit the architectural details in the main paper and instead include them in the supplementary materials.

\subsection{Semantic-Preserving Rewriter}
SPR targets to improve another crucial input for text-to-video generation, \emph{i.e.,} the text prompt, while keeping the semantics aligned with the original text.

\noindent\textbf{Reference-Guided Rewriting} aims to provide descriptions to guide the rewriting, allowing the LLMs to ``imagine'' reasonable textual details.
In particular, we first use the Sentence-BERT model \citep{reimers-2019-sentence-bert} to encode the text prompt. Consider a text prompt $\mathcal{C}$, its top-$k$ references are selected using the cosine similarity:
\begin{equation}
  \{S^r_i\}_{1}^k = \argtopk \{\text{sim}(E_t(\mathcal{C}),E_t(S_{i}))|S_{i}\in \mathcal{X}\}.
\end{equation}

Subsequently, we integrate the references $\{S^r_i\}_{1}^k$ and the text prompt $\mathcal{C}$ into a designed instruction template \texttt{IT}, which is fed into a LLM to rewrite the prompt:
\begin{equation}
    \mathcal{C^{\text{r}}}=\text{LLM}(\texttt{IT}\{\{S^r_i\}_{1}^k,\mathcal{C}\}),
\end{equation}

\noindent\textbf{Denoising with Hybrid Semantics.} Reference sentences provide valuable guidance for the reasonable details compensation, however, we found this strategy cannot perfectly maintain the original semantics of the text prompt due to the excellent association ability of LLMs. To remedy this issue, we propose to introduce the original text prompt into the denoising process. Specifically, we apply the rewritten sentence as the condition in the early stage to boost the content quality, while the original text prompt is employed in the latter denoising steps to pull the semantics close to the original prompt. As a result, we evolve the video synthesis of Eq.~\ref{inference} as follows:
\begin{equation}
    z_{i-1}=\left\{
    \begin{array}{lll}
    \text{DN}(\epsilon_\text{mix}(v*,\eta),t,\mathcal{C}^r), & \text{if}\;{t=T } \\
    \text{DN}(z_{t},t,\mathcal{C^{\text{r}}}) & \text{if}\;{T-m < t \leq T-1 }\\
    \text{DN}(z_{t},t,\mathcal{C}) & \text{if}\;{0 < t \leq T-m},
    \end{array} \right.
\end{equation}
where $m=\lfloor T\times\gamma \rfloor, \gamma\in (0,1).$ $m$ indicates how many steps of rewritten text is performed, $\lfloor\cdot\rfloor$ is the floor operation. Following the latent diffusion model \citep{rombach2021highresolution}, we finally synthesize the video by feeding the final latent feature $z_0$ into the VQ-VAE decoder: $\hat{v}=\mathcal{D}(z_0)$.

\subsection{Discussion}
\noindent\textbf{ONA \emph{vs} Inversion in Video Editing.} Video Editing (VE) approaches often invert the video for editing to initialize the noise ~\citep{wu2022tune, liu2023videop2p, qi2023fatezero}. ONA differs from VE methods in many aspects. \emph{(1) Different goals.} VE inverts the video to preserve more information about the original video, thereby better supporting the subsequent editing.
While ONA uses the inversion to approach the potential optimal noise, which is the core idea of this paper.
\emph{(2) Video diversity.} Video Editing (VE) does not require much consideration of diversity. While our focus is on the general text-to-video generation, consequently, ONA has to maintain the diversity, which is accomplished via the Gaussian mixture.
\emph{(3) Inversion source.} The video for editing is the straightforward inversion source in VE. In contrast, the video for the optimal noise is not available during inference, therefore, ONA resorts to searching  a neighbor video as the inversion source.    


\noindent\textbf{SPR \emph{vs} Text Prompt-Enhancement.} 
Prior research has employed LLMs to enhance text prompts for improving text-to-vision generation~\citep{zhu2023moviefactory,hao2022optimizing, hong2023large, openai2023dalle3}. However, 
there approaches either introduce extra training burden~\cite{hao2022optimizing}, or simply utilize the frozen LLMs to provide additional textual details~\citep{openai2023dalle3}, but rarely study how to better unlock the potential of LLMs when rewriting and pay rare attention to the alignment between the original text and the videos condition on the rewritten text~\cite{zhu2023moviefactory,hong2023large}. 
While SPR is optimization-free, and RGR in our SPR assists LLMs by showing several sentences as references, and DHS in SPR presents a hybrid denoising strategy to keep the semantics of the final video consistent with the original text.
These features set our SPR apart from existing text prompt-rewriting methods.

\section{Experiment}
\label{others}
\subsection{Standard Evaluation Setup.} During our practice, we found that the evaluation settings in existing text-to-video works are either varied or ambiguous, posing the risk of unfair comparison. To remedy this issue, we present our detailed evaluation configuration and hope to standardize the future evaluation of text-to-video models. 

\noindent\textbf{Datasets \&  Metrics.} We follow the previous works \citep{ho2022video, singer2022make, ge2023preserve, blattmann2023align, wang2023videofactory} and adopt the widely-used \textbf{MSR-VTT} \citep{xu2016msr} and \textbf{UCF101} \citep{soomro2012ucf101} datasets for performance evaluation.
\emph{MSR-VTT} provides 2,990 video clips for testing, each accompanied by around 20 captions. To evaluate the performance, we randomly select one caption per video to create a set of 2,990 text-video evaluation pairs. FID \citep{heusel2017gans}\footnote{FID: https://github.com/GaParmar/clean-fid} and CLIP-FID \citep{kynkaanniemi2022role}\footnote{CLIP-FID: https://github.com/GaParmar/clean-fid} is used to assess the video quality, along with CLIP-SIM \citep{wu2021godiva}\footnote{CLIPSIM: https://github.com/openai/CLIP} metric to measure semantic consistency between videos and text prompts. 
\emph{UCF101} contains 13,320 video clips of 101 human action categories. We evaluate performance using 3,783 test videos. As there are no captions, we use the text prompts from PYoCo \citep{ge2023preserve} for video generation. IS \citep{salimans2016improved} and FVD \citep{unterthiner2018towards, yan2021videogpt} serve as the evaluation criteria. Video IS~\citep{singer2022make,hong2022cogvideo} uses C3D\citep{tran2015learning} pretrained on UCF101 as the feature extractor\footnote{IS: https://github.com/pfnet-research/tgan2}, while FVD utilizes I3D \citep{vadisaction} pretrained on Kinetics-400~\citep{kay2017kinetics} for video feature encoding\footnote{FVD: https://github.com/wilson1yan/VideoGPT}.

\noindent\textbf{Resolution.} Following previous works \citep{singer2022make,zhou2022magicvideo, ge2023preserve, he2022latent}, we generate videos of size $16 \times 256 \times 256$ for performance evaluation. This choice is motivated by two main factors: First, the training dataset WebVid-10M \citep{bain2021frozen} for most text-to-video models is dominated by 360P videos; Second, both the MSR-VTT and UCF101 datasets consist of videos with a resolution of around 240P. Consequently, $256\times 256$ ensures a consistent and comparable evaluation setting. 

\newcolumntype{S}{@{}>{\lrbox0}l<{\endlrbox}}  %
\definecolor{lightgreen}{HTML}{D8ECD1}
\newcommand{\better}[1]{\colorbox{lightgreen}{#1}}

\begin{table*}[t]
\vspace{-0.2cm}
\tablestyle{2pt}{1.1}
\begin{tabular}{*l| ^c ^c ^c ^c| ^c ^c ^c}
& 
\multicolumn{4}{c|}{\textbf{MSR-VTT}} &
\multicolumn{3}{c}{\textbf{UCF101}} \\
&
\multicolumn{1}{c|}{FID (inception v3)↓} & 
\multicolumn{1}{c|}{CLIP-FID ↓} &
\multicolumn{1}{c|}{ CLIPSIM ↑} &
\multicolumn{1}{c|}{Hyperparamters} &
\multicolumn{1}{c|}{FVD ↓} &
\multicolumn{1}{c|}{ IS ↑}  &
 Hyperparameters \\
\shline

$\text{Tune-A-Video}^{\dag}$  & 
\hspace{-0.765cm} 51.365&\hspace{-0.755cm} 17.093& \hspace{-0.74cm} 0.276 & - &
\hspace{-0.75cm} 1321.45 & \hspace{-0.75cm} 22.752 & - \\

{\qquad + ONA} & 
\hspace{-0.05cm}${48.623}_{\hi{2.743}}$& ${16.393}_{\hi{0.700}}$& \hspace{-0.05cm}$\textbf{0.278}_{\hi{0.002}}$ & $\eta=0.1$ &
\hspace{0.15cm}${1156.91}_{\hi{164.54}}$ & ${25.129}_{\hi{2.377}}$ & $\eta=0.1$  \\

{\qquad + SPR} & 
\hspace{-0.07cm}${47.389}_{\hi{3.976}}$& ${16.436}_{\hi{0.657}}$& \hspace{0.2cm}${0.273}_{\hie{-0.003}}$ & $\gamma=0.5$ &
\hspace{-0.05cm} ${1276.67}_{\hi{44.78}}$ & ${24.960}_{\hi{2.208}}$ & $\gamma=0.5$ \\

{\qquad + POS} & 
\hspace{-0.12cm}${\textbf{45.270}}_{\hi{6.095}}$& ${\textbf{16.164}}_{\hi{0.929}}$& \hspace{0.20cm}${0.274}_{\hie{-0.002}}$ & $\{\eta=0.1,\gamma=0.5\}$  &
\hspace{0.1cm}${\textbf{1078.54}}_{\hi{242.91}}$ & $\textbf{27.317}_{\hi{4.565}}$ &  $\{\eta=0.1,\gamma=0.5\}$\\
\shline

VideoCrafter  & 
\hspace{-0.765cm} 51.349&\hspace{-0.755cm} 20.899& \hspace{-0.74cm} 0.282 & - &
\hspace{-0.82cm} 777.02 & \hspace{-0.75cm} 31.211 & -\\

{\qquad + ONA} & 
\hspace{0.17cm}${51.488}_{\hie{-0.139}}$& \hspace{-0.06cm}${\textbf{20.808}}_{\hi{0.091}}$& \hspace{-0.05cm}${\textbf{0.285}}_{\hi{0.003}}$ & $\eta=0.3$ &
\hspace{-0.1cm}${764.86}_{\hi{12.16}}$ & \hspace{-0.08cm}$\textbf{32.439}_{\hi{1.228}}$ & $\eta=0.3$ \\

{\qquad + SPR} & 
\hspace{-0.1cm}${\textbf{50.221}}_{\hi{1.128}}$& ${20.870}_{\hi{0.029}}$& ${0.283}_{\hi{0.001}}$ & $\gamma=0.1$ &
\hspace{-0.2cm} ${\textbf{746.91}}_{\hi{30.11}}$ & ${31.242}_{\hi{0.031}}$ & $\gamma=0.1$ \\

{\qquad + POS} & 
\hspace{-0.08cm}${50.882}_{\hi{0.467}}$& ${20.886}_{\hi{0.013}}$& ${0.284}_{\hi{0.002}}$ & $\{\eta=0.3,\gamma=0.1\}$ &
\hspace{-0.15cm} ${747.32}_{\hi{29.70}}$ & ${32.372}_{\hi{1.161}}$ & $\{\eta=0.3,\gamma=0.1\}$ \\
 
\shline

ModelScope & 
\hspace{-0.765cm} 45.378&\hspace{-0.755cm} 13.677& \hspace{-0.74cm} 0.296 & - &
\hspace{-0.82cm} 774.14 & \hspace{-0.75cm} 32.337 \\

{\qquad + ONA} & 
\hspace{-0.08cm}${43.092}_{\hi{2.287}}$& \hspace{-0.08cm}${\textbf{13.572}}_{\hi{0.105}}$& \hspace{-0.03cm}$\text{0.299}_{\hi{0.003}}$ &  $\eta=0.2$ &
${607.11}_{\hi{167.03}}$ & \hspace{-0.08cm}$\textbf{38.992}_{\hi{6.655}}$ &  $\eta=0.2$ \\

{\qquad + SPR} & 
\hspace{-0.08cm}${43.585}_{\hi{1.793}}$& \hspace{-0.08cm}${\textbf{13.572}}_{\hi{0.105}}$& ${0.297}_{\hi{0.001}}$ &  $\gamma=0.04$ &
\hspace{-0.2cm} ${684.90}_{\hi{89.24}}$ & ${33.136}_{\hi{0.799}}$ &  $\gamma=1.0$ \\

{\qquad + POS} & 
\hspace{-0.1cm}${\textbf{42.755}}_{\hi{2.623}}$& \hspace{-0.08cm}${13.674}_{\hi{0.003}}$& ${\textbf{0.300}}_{\hi{0.004}}$ & $\{\eta=0.2,\gamma=0.04\}$ &
\hspace{-0.1cm} ${\textbf{566.68}}_{\hi{207.46}}$ & ${38.190}_{\hi{5.853}}$ & $\{\eta=0.2,\gamma=1.0\}$ \\
\hline

SCVideo & 
\hspace{-0.765cm} 48.331& \hspace{-0.765cm} 14.802 & \hspace{-0.765cm} 0.283 & - &
\hspace{-0.82cm} 750.74 & \hspace{-0.765cm} 23.224 & - \\

{\qquad + ONA} & 
\hspace{-0.09cm}$43.585_{\hi{4.746}}$& \hspace{-0.08cm}${14.287}_{\hi{0.515}}$ & \hspace{-0.03cm}${\textbf{0.289}}_{\hi{0.006}}$ & $\eta=0.5$ &
${558.51}_{\hi{192.23}}$ & ${31.538}_{\hi{8.314}}$ & $\eta=0.5$ \\

{\qquad + SPR} & 
\hspace{-0.08cm}${43.760}_{\hi{4.571}}$ & \hspace{-0.08cm}${14.291}_{\hi{0.511}}$& ${0.284}_{\hi{0.001}}$ & $\gamma=0.1$ &
\hspace{-0.25cm} $\text{712.88}_{\hi{37.86}}$ & ${25.648}_{\hi{2.424}}$ & $\gamma=1.0$ \\

{\qquad + POS} & 
\hspace{-0.1cm}${\textbf{42.902}}_{\hi{5.429}}$& \hspace{-0.1cm}${\textbf{14.155}}_{\hi{0.647}}$& ${0.288}_{\hi{0.005}}$ & $\{\eta=0.5,\gamma=0.1\}$ &
\hspace{-0.1cm} ${\textbf{541.43}}_{\hi{209.31}}$ & $ \hspace{0.1cm} {\textbf{33.245}}_{\hi{10.021}}$  & $\{\eta=0.5,\gamma=1.0\}$ \\

\end{tabular}

\caption{\textbf{Quantitative results} on MSR-VTT and UFC101 datasets, our POS, and its components ONA and SPR can improve the performance with a clear margin, where  ``+ ONA'' means equipping ONA to the models, and \textbf{Bold} highlights the best performance. \hi{Green} and \hie{Red} numbers means performance improvement and decline, respectively}.
\vspace{-0.3cm}
\label{tab:method}

\end{table*}

\begin{table*}[t]
\tablestyle{2pt}{1.1}

\resizebox{\linewidth}{!}{

\begin{tabular}{*l| ^c ^c ^c ^c| ^c ^c ^c}
& 
\multicolumn{4}{c|}{\textbf{MSR-VTT}} &
\multicolumn{3}{c}{\textbf{UCF101}} \\
&
\multicolumn{1}{c|}{FID (inception v3)↓} & 
\multicolumn{1}{c|}{CLIP-FID ↓} &
\multicolumn{1}{c|}{ CLIPSIM ↑} &
\multicolumn{1}{c|}{Hyperparamters} &
\multicolumn{1}{c|}{FVD ↓} &
\multicolumn{1}{c|}{ IS ↑}  &
 Hyperparameters \\
\shline

ModelScope & 
\hspace{-0.765cm} 45.378&\hspace{-0.755cm} 13.677& \hspace{-0.74cm} 0.296 & - &
\hspace{-0.82cm} 774.14 & \hspace{-0.75cm} 32.337 \\

{\qquad + POS} & 
\hspace{-0.1cm}${42.755}_{\hi{2.623}}$& \hspace{-0.08cm}${\textbf{13.674}}_{\hi{0.003}}$& ${\textbf{0.300}}_{\hi{0.004}}$ & $\{\eta=0.2,\gamma=0.04\}$ &
\hspace{-0.1cm} ${\textbf{566.68}}_{\hi{207.46}}$ & ${38.190}_{\hi{5.853}}$ & $\{\eta=0.2,\gamma=1.0\}$ \\

{\qquad + POS${^{*}}$} & 
\hspace{-0.1cm}${\textbf{42.296}}_{\hi{3.082}}$& \hspace{-0.08cm}${13.881}_{\hie{-0.204}}$& ${0.299}_{\hi{0.003}}$ & $\{\eta=0.2,\gamma=0.1\}$ &
\hspace{-0.1cm} ${570.27}_{\hi{203.87}}$ & ${\textbf{44.255}}_{\hi{11.918}}$ & $\{\eta=0.5,\gamma=0.1\}$ \\
\hline

SCVideo & 
\hspace{-0.765cm} 48.331& \hspace{-0.765cm} 14.802 & \hspace{-0.765cm} 0.283 & - &
\hspace{-0.82cm} 750.74 & \hspace{-0.765cm} 23.224 & - \\

{\qquad + POS} & 
\hspace{-0.1cm}${42.902}_{\hi{5.429}}$& \hspace{-0.1cm}${\textbf{14.155}}_{\hi{0.647}}$& ${0.288}_{\hi{0.005}}$ & $\{\eta=0.5,\gamma=0.1\}$ &
\hspace{-0.1cm} ${\textbf{541.43}}_{\hi{209.31}}$ & $ \hspace{0.1cm} {33.245}_{\hi{10.021}}$  & $\{\eta=0.5,\gamma=1.0\}$ \\

{\qquad + POS${^{*}}$} & 
\hspace{-0.1cm}${\textbf{42.838}}_{\hi{5.493}}$& \hspace{-0.1cm}${14.517}_{\hi{0.285}}$& ${\textbf{0.292}}_{\hi{0.009}}$ & $\{\eta=0.5,\gamma=0.1\}$ &
\hspace{-0.1cm} ${543.73}_{\hi{207.01}}$ & $ \hspace{0.1cm} {\textbf{38.575}}_{\hi{15.31}}$  & $\{\eta=0.5,\gamma=0.5\}$ \\

\end{tabular}}
\caption{\textbf{Quantitative results} on MSR-VTT and UFC101 datasets, our POS${^{*}}$ is capable of achieving a performance similar to that of POS, with both significantly outperforming the baseline model.}
\label{tab:method_1}
\vspace{-0.6cm}
\end{table*}

\begin{figure*}
    \centering
    \hspace{-0.1cm}\includegraphics[height=9cm]{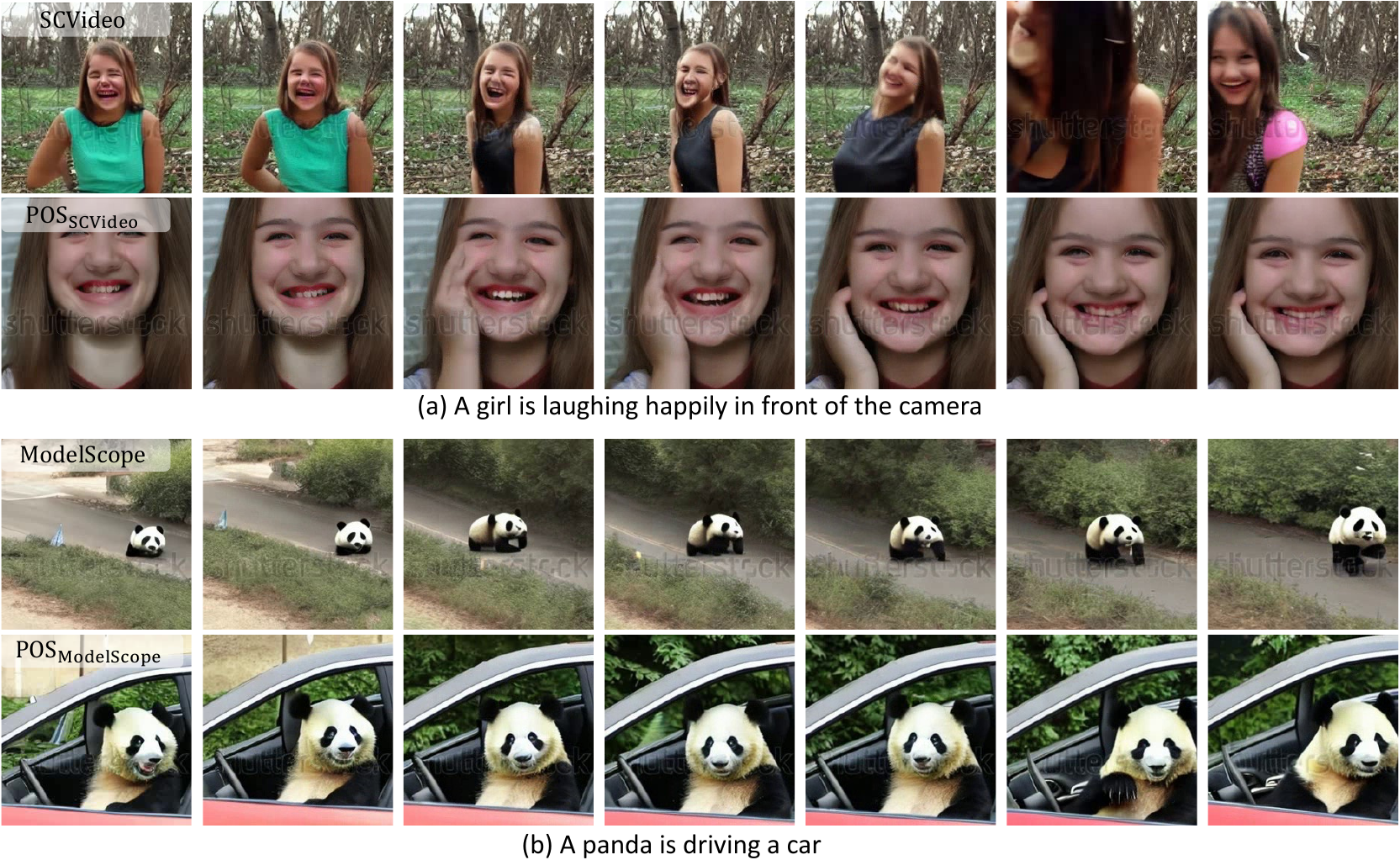}
    \vspace{-0.2cm}
    \caption{\textbf{Qualitative results.} $\text{POS}_{\text{ModelScope}}$ means ModelScope with POS equipped, subfigures (a) and (b) show the results with SCVideo and ModelScope as backbones. Each group shares the same random noise for a fair comparison. }
    \label{fig:qualitativeResult}
\end{figure*}

\noindent\textbf{Sampling Strategy.} For FID and CLIP-FID, we sample 14,950 images from 2,990 MSR-VTT videos and generated videos respectively. Concretely, we sample 5 images for each video in both real and generated videos. For the real video, sample an image every 12 frames. For the generated video, sample an image every 4 frames. Regarding CLIP-SMI, we calculate the clip similarity between each image of the generated video and the corresponding prompt and then average it. All 3,783 UCF101 test videos are used to calculate FVD and IS. For the real video, we sample an image every 5 frames, for a total of 16 frames, to calculate FVD.

\subsection{Implementation Details}
We sample 100k image-text pairs from WebVid-10M to form our pre-defined candidate pool, \emph{i.e.,} $N=100$k. In Guided Noise Inversion, we search for the most similar video from the pool for each text prompt according to Eq.~\ref{eq6}. In Noise Prediction Network, we randomly sample 600K video-text pairs from WebVid-10M and utilize DDIM inversion to convert the videos to noise. More inverted videos  harvests limited benefits in our practice. Sentence-BERT is taken to extract text features\footnote{Sentence-BERT: https://huggingface.co/sentence-transformers/paraphrase-multilingual-MiniLM-L12-v2}.
In Semantic-Preserving Rewriter, we pick the top-5 ($k=5$) reference sentences for the text prompt and adopt ChatGPT \cite{openai2023gpt4} to perform rewriting, which imitates the adjectives, adverbs, or sentence patterns of these 5 sentences. The instruction template \texttt{IT} is  \emph{``Let me give you 5 examples:}\texttt{[Ref1],[Ref2],[Ref3],[Ref4],[Ref5]}, \emph{rewrite the sentence} \texttt{[Input Text Prompt]} \emph{without changing the meaning of the original sentence to a maximum of 20 words, imitating/combining the adjectives, adverbs or sentence patterns from the 5 examples above''}. 
POS can benefit many trained text-to-video models. To support this claim, we equip the proposed modules on three open-sourced text2video models, including ModelScope \cite{wang2023modelscope}, VideoCrafter \cite{he2022lvdm}, Tune-A-Video \cite{wu2022tune} and our trained SCVideo to study the performance gain. Tune-A-Video is a well-known one-shot video generation model, to perform the general video generation, we re-train this model by optimizing more parameters of spatiotemporal attention and 3D convolution module on WebVid-10M by referring to many text-to-video works\cite{blattmann2023align, ge2023preserve, he2022latent}. The re-trained Tune-A-Video is marked as $\text{Tune-A-Video}^{\dag}$. SCVideo is our designed model extended from text-to-image models by equipping temporal modules, including temporal convolution, Sparse Causal (SC) Attention, and temporal attention.

\noindent\textbf{Training.} Both SCVideo and the Noise Prediction Network employ the same UNet architecture and identical loss functions. The difference lies in the initialization of network parameters. SCVideo is initialized using the pretrained weights of SD 1.4, and the newly introduced weights and bias are initialized with identity matrix and zero, respectively. NPNet is initialized from the parameters of a well-trained SCVideo.  Additionally, we only optimize the 3D temporal convolution and all the attention modules of the former, whereas the latter undergoes training for all parameters. SCVideo is trained with 4 Nvidia A100-80G GPUS with batch size 5, and the total number of training steps per GPU is 80K. The learning rate is set as $3\times 10^{-5}$ with $10^{-2}$weight decay. We train our model using AdamW optimizer, and the hyperparameters are configured as  $\beta_1$ = 0.9, $\beta_2$ = 0.999 and $\epsilon$ = $10^{-8}$. In addition, a cosine schedule with a linear warmup of 1000 steps is adopted during the training process. NPNet is trained with 2 Nvidia A100-80G GPUS with batch size 5, and the total number of training steps per GPU is 40K. Other hyperparameter settings are the same as SCVideo.

Next, we extensively evaluate our approach on multiple baseline models. ONA and POS denote the use of retrieved videos and DDIM Inversion, while POS$^{*}$ represents the utilization of Noise Prediction Network

\begin{table*}[t]
\vspace{-0.2cm}
\tablestyle{4pt}{1.1}
\begin{tabular}{*l| ^c ^c ^c ^c| ^c ^c ^c ^c}
\multirow{2}*{\diagbox{Metrics}{Backbones}}& 
\multicolumn{4}{c|}{\textbf{Stable Diffusion 1.5}} & 
\multicolumn{4}{c}{\textbf{Stable Diffusion-XL}} \\
&
\multicolumn{1}{c|}{\scriptsize SD 1.5} & 
\multicolumn{1}{c|}{\scriptsize + ONA} &
\multicolumn{1}{c|}{\scriptsize + SPR} & 
\multicolumn{1}{c|}{\scriptsize + POS} &
\multicolumn{1}{c|}{\scriptsize SD-XL} & 
\multicolumn{1}{c|}{\scriptsize + ONA} &
\multicolumn{1}{c|}{\scriptsize + SPR} & 
\multicolumn{1}{c}{\scriptsize + POS}\\
\shline
FID (inception v3)↓ & 
24.824& ${24.641}_{\hi{0.183}}$& ${23.861}_{\hi{0.963}}$ &
${\textbf{23.623}}_{\hi{1.201}}$
& 25.608 & ${24.129}_{\hi{1.479}}$ & ${25.307}_{\hi{0.301}}$ & ${\textbf{24.011}}_{\hi{1.597}}$
\\

CLIP-FID ↓ & 
${13.545}$ & ${13.352}_{\hi{0.193}}$ & ${13.212}_{\hi{0.333}}$ &
${\textbf{13.194}}_{\hi{0.351}}$
& 14.940& ${12.885}_{\hi{2.055}}$& ${14.717}_{\hi{0.223}}$ & ${\textbf{12.866}}_{\hi{2.074}}$
\\
CLIPSIM ↑ & 
\hspace{-0.11cm}0.324& $ \hspace{-0.1cm} {\textbf{0.325}}_{\hi{0.001}}$ & \hspace{0.03cm} ${0.320}_{\hie{-0.004}}$ & \hspace{0.06cm}
${0.320}_{\hie{-0.004}}$
& \hspace{-0.03cm}0.336 & \hspace{-0.23cm} ${\textbf{0.337}}_{\hi{0.001}}$ & \hspace{0.01cm} ${0.330}_{\hie{-0.006}}$ & \hspace{0.05cm} ${0.331}_{\hie{-0.005}}$
\\
\end{tabular}
\vspace{-0.2cm}
\caption{\textbf{Quantitative results} for text-to-image generation, the results are evaluated on MS-COCO dataset}.
\label{tab:image evaluation}
\vspace{0.2cm}
\end{table*}

\begin{table*}[h]
\vspace{-1em}
\subfloat[Size of candidate pool.
\label{tab:poolDis}
]{
   \hspace{-0.8em} \begin{minipage}[r]
        {0.45\linewidth}{
            \begin{center}
            \vspace{-0.1cm}
            \tablestyle{4pt}{1.1}\scriptsize
\begin{tabular}{*l| ^c ^c ^c| ^c ^c}
& 
\multicolumn{3}{c|}{\textbf{MSR-VTT}} &
\multicolumn{2}{c}{\textbf{UCF101}} \\
&
\multicolumn{1}{c|}{\scriptsize FID↓} & 
\multicolumn{1}{c|}{\scriptsize CLIP-FID↓} &
\multicolumn{1}{c|}{\scriptsize CLIPSIM↑} &
\multicolumn{1}{c|}{\scriptsize FVD↓} &
\scriptsize IS ↑ \\
\shline

{$N$= 10K} & 
43.236& 14.352& 0.288 &
591.79 & 31.735  \\

{$N$= 100K} & 
43.585& 14.287& \textbf{0.289} &
\textbf{558.51} & 31.538  \\

{$N$= 1M} & 
42.587& 14.334& 0.288 &
620.29 & 33.047  \\

{$N$= 10M} & 
\textbf{42.072}& \textbf{14.285}& \textbf{0.289} &
571.27 & \textbf{33.232}  
\vspace{0.42cm}

\end{tabular}
            \end{center}
        }
    \end{minipage}
}
\subfloat[Proportion of Rewritten Text in DHS.
\label{tab:diffStep}
]{
    \centering
    \begin{minipage}[l]
        {0.45\linewidth}{
            \begin{center}
            \vspace{-0.1cm}
            \tablestyle{4pt}{1.1}\scriptsize
\begin{tabular}{*l| ^c ^c ^c| ^c ^c}
& 
\multicolumn{3}{c|}{\textbf{MSR-VTT}} &
\multicolumn{2}{c}{\textbf{UCF101}} \\
&
\multicolumn{1}{c|}{\scriptsize FID↓} & 
\multicolumn{1}{c|}{\scriptsize CLIP-FID↓} &
\multicolumn{1}{c|}{\scriptsize CLIPSIM↑} &
\multicolumn{1}{c|}{\scriptsize FVD↓} &
\scriptsize IS↑ \\
\shline

{$\gamma$=0} & 
48.331 & 14.802 & 0.283 &
750.74 & 23.224  \\

{$\gamma$=0.1} & 
43.760& \textbf{14.291}& \textbf{0.284} &
769.35 & 25.006  \\

{$\gamma$=0.2} & 
43.170& 14.384& 0.283 &
731.48 & 25.508 \\

{$\gamma$=0.5} & 
\textbf{42.992}& 14.755& 0.278 &
729.23 & 25.453 \\

{$\gamma$=1.0} & 43.531 & 15.484 & 0.275 &
\textbf{712.88} & \textbf{25.648}
\vspace{0.1cm}

\end{tabular}
            \end{center}
        }
    \end{minipage}
}
\\
\hspace{-2em}
\subfloat[Number of reference text in SPR.
\label{tab:numTextDis}
]{
    \begin{minipage}
    {0.45\linewidth}{
        \begin{center}
        \vspace{-0.1cm}
        \tablestyle{4pt}{1.1}\scriptsize
\begin{tabular}{*l| ^c ^c ^c| ^c ^c}
& 
\multicolumn{3}{c|}{\textbf{MSR-VTT}} &
\multicolumn{2}{c}{\textbf{UCF101}} \\
&
\multicolumn{1}{c|}{\scriptsize FID↓} & 
\multicolumn{1}{c|}{\scriptsize CLIP-FID↓} &
\multicolumn{1}{c|}{\scriptsize CLIPSIM↑} &
\multicolumn{1}{c|}{\scriptsize FVD↓} &
\scriptsize IS↑ \\
\shline

{K=0} & 
45.435& 15.504& \textbf{0.278} &
808.16 & 23.849  \\

{K=1} & 
45.410& 15.386& 0.273 &
745.07 & 24.604  \\

{K=2} & 
44.458& 15.385& 0.275 &
824.99 & 25.361  \\

{K=5} & 
43.531& 15.484& 0.275 &
\textbf{712.88} & 25.648  \\

{K=10} & 
\textbf{43.125}& \textbf{15.257}& 0.277 &
755.55 & \textbf{26.598}  \\

\end{tabular}
        \end{center}
    }
    \end{minipage}
}
\hspace{-2em}
\centering
\subfloat[Performance of different rewriting engines.
\label{tab:llms}
]{
    \centering
    \begin{minipage}
    {0.50\linewidth}{
    \vspace{-0.1cm}
        \begin{center}
        \tablestyle{4pt}{1.1}\scriptsize
\begin{tabular}{*l| ^c ^c ^c| ^c ^c}
& 
\multicolumn{3}{c|}{\textbf{MSR-VTT}} &
\multicolumn{2}{c}{\textbf{UCF101}} \\
&
\multicolumn{1}{c|}{\scriptsize FID↓} & 
\multicolumn{1}{c|}{\scriptsize CLIP-FID↓} &
\multicolumn{1}{c|}{\scriptsize CLIPSIM↑} &
\multicolumn{1}{c|}{\scriptsize FVD↓} &
\scriptsize IS↑ \\
\shline
{Llama2-7B} & 
43.811 & 14.395 & 0.287 &
551.87 & 31.305  \\
{ChatGPT} & 
\textbf{42.901} & \textbf{14.155} & \textbf{0.288} &
\textbf{541.43} & \textbf{33.245}  \\
\end{tabular}
        \end{center}
    }
    \vspace{0.576cm}
    \end{minipage}
}
\vspace{-0.2cm}
\caption{\textbf{Ablation experiments}. SCVideo is taken as the baseline to study the key hyperparameters in our ONA and SPR. 
}
\label{tab:test}
\end{table*}

\begin{figure*}
    \centering
    \includegraphics[width=14cm,height=3.3cm]{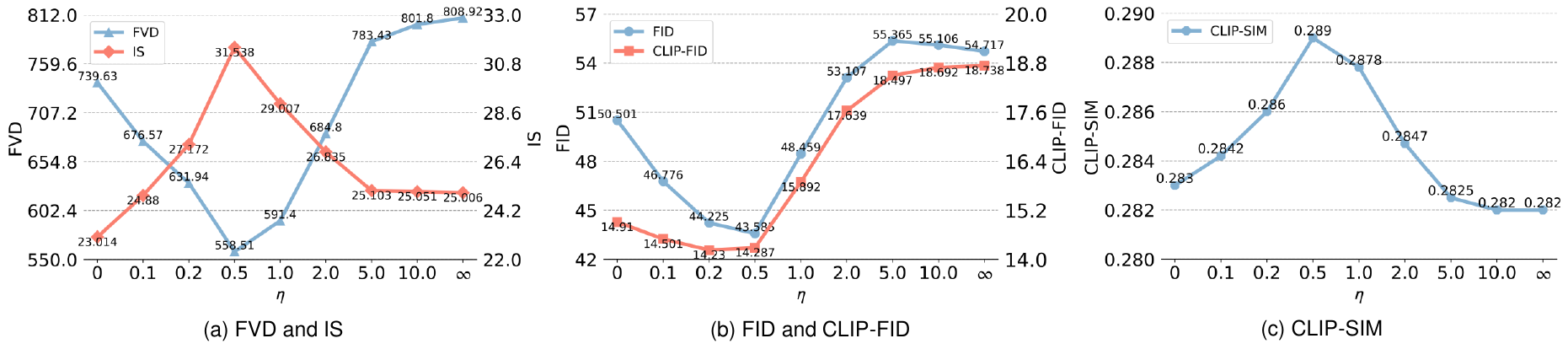}
    \vspace{-0.2cm}
    \caption{\textbf{Ablation study} on the hyperparameter $\eta$ in ONA, 
    $\eta=0.5$ achieves the best performance (SCVideo serves as the baseline).}
    \vspace{-0.4cm}
    \label{fig:inversion}
\end{figure*}

\subsection{Quantitative Results.}
Table \ref{tab:method} reports the main results on four benchmarks, the hyperparameter configurations are also presented to make everything clear. We can observe that the POS can bring consistent performance improvement of four models on both datasets. Particularly, ONA and SPR can improve the FID of ModelScope from 45.378 to 43.092 and 43.585, respectively. Equipping both modules (POS) makes the performance step further, achieving a 42.755 FID. Furthermore, POS also maintains a good semantic consistency, 0.296 \emph{vs} 0.3. On UCF101 dataset, ONA shows great effectiveness, boosts the FVD of ModelScope from 774.14 to 607.11, and IS is also clearly improved.  The contributions on our SCVideo are more remarkable, we can harvest an improvement of 5.429 FID on MSR-VTT and 209.31 of FVD on UCF101. $\text{Tune-A-Video}^{\dag}$ is also clearly improved by our POS, whose FID and FVD can be boosted by a margin of 6.095 and 242.91, respectively.  In comparison, VideoCrafter only harvests incremental performance gains from ONA, the reason mainly stems from the poor inversion quality. In our practice, we found that the noise from VideoCrafter's inversion function cannot well reconstruct the video, which indicates that the inverted noise fails to locate the optimal noise, thereby providing a not good guidance for ONA. Figure~\ref{fig:qualitativeResult} shows two groups of qualitative results from two SOTA models, ModelScope and SCVideo, from which we can intuitively observe the effectiveness of our POS. 

As shown in Table \ref{tab:method_1}, with the assistance of NPNet, POS* significantly enhances the performance of the baseline models, achieving comparable results to POS utilizing the retrieval pool. This provides us with two options: 1) Employing the training-free POS with a retrieval pool. 2) Leveraging POS$^{*}$ with a pre-trained NPNet to overcome storage issues.

\subsection{Ablations}
\noindent\textbf{Can POS Benefit Image Generation?} To answer this question, we augment the Stable Diffusion (SD) v1.5 \citep{rombach2021highresolution} and Stable Diffusion-XL (SD-XL) \citep{podell2023sdxl} with our POS and evaluate their performance on MS-COCO test set \citep{lin2014microsoft}.  We employ the test set of Flickr30k \citep{young2014image} as the candidate pool instead of the MS-COCO training set to verify the generalization of POS, and $\eta$ and $\gamma$ are set as 0.05 and 0.4, the other hyperparameters remain the same as video experiments. 

Table \ref{tab:image evaluation} compares the results, we have the following observations: (a) ONA yields a more substantial performance improvement for SD-XL compared to SD 1.5. This discrepancy arises from the fact that SD-XL samples noise from a larger space (128×128) in comparison to SD 1.5 (64×64), thereby rendering the task of achieving or approximating optimal noise more challenging for SD-XL. Consequently, ONA can leverage its strengths more effectively in the case of SD-XL. Conversely, the noise space of SD 1.5 is relatively constrained, and the extensive training with a voluminous dataset has effectively aligned the noise and image spaces. Consequently, the efficacy of ONA is less pronounced in this context. (b) SPR exhibits superior performance on SD 1.5 in contrast to SD-XL. This phenomenon can be attributed to SD-XL training a more potent text encoder, which is adept at capturing finer textual details.

\noindent\textbf{Size of Candidate Pool.} 
To investigate the effect of candidate pool size, we randomly sample 10K, 100K, 1M, and 10M samples from the Webvid dataset to study the performance trend. 
From Table \ref{tab:poolDis},  we can observe that the larger the retrieval pool typically results in better performance.  For example, enlarging the pool size from 10K to 10M can promote the FID on MSR-VTT from 43.236 to 42.072. Notably, although our default set is $N=$100K,  reducing the scale to 10K does not severely hinder the performance. We take 100K to pursue a better trade-off between the scale of the candidate pool and the performance.

\noindent\textbf{Proportion of Guided Noise in ONA.}
Hyperparameter $\eta$ in Eq.~\ref{eq7} determines the final composition of the noise, this part discusses the performance impact of how to hybridze the two types of noises by varying $\eta$. The results are reported in Figure \ref{fig:inversion},
We can observe a clear tendency from the figure, that using the pure random noise ($\eta=0$) or guided noise ($\eta=\infty$) can not yield satisfactory performance. Instead, an appropriate fusion of the two noise types is a more effective manner, we can harvest the best performance when $\eta=0.5$. 


\noindent\textbf{Number of Dominated Steps of Rewritten Text in DHS.} 
The hyperparameter $\gamma$ controls how many steps of the rewritten text are performed in DHS. We vary $\gamma$ in this part and study its effect on performance, and do not equip the ONA to observe a clearer trend. 
As shown in Table \ref{tab:diffStep},  the rewritten sentence ($\gamma=1$) can boost the quality compared to the raw prompt. 
Specifically, compared with the baseline($\gamma=0$ without rewritten text), applying the rewritten sentence reduces FID from 48.331 to 43.531 and promotes the IS score from 23.224 to 25.648. However, the CLIP-SIM drops from 0.283 to 0.275, indicating the semantic consistency is deteriorated. Our DHS strategy can remedy the issue, for example, introducing the original text in the last 90\% denoising steps ($\gamma=0.1$) can boost the CLIP-SIM from 0.275 to 0.284. 

\noindent\textbf{Number of Reference Text for Rewriting.} 
By default, we employ 5 reference sentences for the text rewriting, this subsection examines the impact of varying the number of references for text rewriting. Table~\ref{tab:numTextDis} presents the performance results across different reference numbers, highlighting the positive impact of incorporating references (we focus solely on the rewriting strategy, excluding the ONA and DHS techniques). For instance, the FID score is significantly improved from 45.435 without references to 43.531 with five references. Despite consistent improvements in other metrics, CLIP-SIM shows a decline due to the introduction of new content or objects. However, this issue can be effectively mitigated by our DHS mechanism, as evidenced in Table~\ref{tab:method} and Table~\ref{tab:diffStep}.

\noindent\textbf{Effect of Large Language Models (LLMs).} ChatGPT is taken as the rewriting engine due to its excellent performance, we also present a discussion regarding the LLMs to further validate our methods.
Table~\ref{tab:llms} compares the results of using ChatGPT (GPT-3.5-Turbo) and Llama2-7B in SPR. 
The findings indicate that ChatGPT consistently outperforms Llama2-7B across five key metrics, exhibiting a substantial advantage. 
Moreover, ChatGPT also shows a higher level of intelligence in our practice, as it does not need regularization to filter out irrelevant content.

\section{Conclusion}
This work presents POS, a model-agnostic suite for enhancing diffusion-based text-to-video generation by improving two crucial inputs: the noise and the text prompt. To approximate the optimal noise for a given text prompt, we propose an optimal noise approximator. This module involves a two-stage process, starting with the search for a video neighbor closely related to the text prompt, and subsequently performing DDIM inversion on the selected video. Alternatively, we also present another solution: training a Noise Prediction Network to overcome storage and search issues associated with the candidate pool, while achieving comparable performance.
Additionally, we devise a semantic-preserving rewriter to enrich the details in the text prompt, aiming to augment the original text input by providing more comprehensive information. To allow a reasonable detail compensation and maintain semantic consistency, we propose a reference-guided rewriting approach and incorporate hybrid semantics during the denoising stage for semantics preserving.
To evaluate the effectiveness of our method, we integrate the proposed POS into four backbones and conduct extensive experiments using the widely used benchmarks. The experimental results demonstrate the efficacy of our approach in enhancing text-to-video models.

\clearpage 

{
    \small
    \bibliographystyle{ieeenat_fullname}
    \bibliography{main}
}


\end{document}